\documentclass[11pt, a4paper, copyright, nonumbering]{klear}
\usepackage[compress, round]{natbib}
\usepackage{dblfloatfix}
\usepackage{caption}
\usepackage{dramatist}
\usepackage{xspace}
\usepackage{pifont}
\usepackage{multirow}
\usepackage{tcolorbox}
\usepackage{xltabular}
\usepackage{longtable}
\usepackage{hyperref}
\usepackage{amsfonts}
\usepackage{amsmath}
\usepackage{amssymb}
\usepackage{lineno}

\usepackage[bottom]{footmisc}

\usepackage{CJKutf8}
\usepackage{subfigure}
\usepackage{setspace}

\definecolor{keywordcolor}{rgb}{0.7, 0.1, 0.1}   
\definecolor{tacticcolor}{rgb}{0.0, 0.1, 0.6}    
\definecolor{commentcolor}{rgb}{0.4, 0.4, 0.4}   
\definecolor{symbolcolor}{rgb}{0.0, 0.1, 0.6}    
\definecolor{sortcolor}{rgb}{0.1, 0.5, 0.1}      
\definecolor{attributecolor}{rgb}{0.7, 0.1, 0.1} 
\usepackage{listings}

\usepackage{tcolorbox}
\newtcolorbox{promptbox}[1][]{
    colback=gray!10,      
    colframe=black!50,     
    boxrule=0.5mm,        
    arc=1mm,              
    boxsep=0mm,
    fontupper=\ttfamily\scriptsize,  
    width=\textwidth,     
    title=#1,
    fonttitle=\ttfamily\footnotesize\centering,
}

\makeatletter
\def\@BTrule[#1]{%
  \ifx\longtable\undefined
    \let\@BTswitch\@BTnormal
  \else\ifx\hline\LT@hline
    \nobreak
    \let\@BTswitch\@BLTrule
  \else
     \let\@BTswitch\@BTnormal
  \fi\fi
  \global\@thisrulewidth=#1\relax
  \ifnum\@thisruleclass=\tw@\vskip\@aboverulesep\else
  \ifnum\@lastruleclass=\z@\vskip\@aboverulesep\else
  \ifnum\@lastruleclass=\@ne\vskip\doublerulesep\fi\fi\fi
  \@BTswitch}
\makeatother

\addto\extrasenglish{
}

 {\begin{list}{}%
         {\setlength{\leftmargin}{#1}}%
         \item[]%
 }
 {\end{list}}

\newcommand{\kylean}{Leanabell-Prover}
\newcommand{\goedeli}{Goedel-Prover}
\newcommand{\goedelii}{Goedel-Prover-SFT}
\newcommand{\dsi}{DeepSeek-Prover-v1.5}
\newcommand{\dsii}{DeepSeek-Prover-v1.5-SFT}
\newcommand{\dsiii}{DeepSeek-Prover-v1.5-RL}

\title{\centering \kylean: Posttraining Scaling in Formal Reasoning}

\author[*]{
\quad \quad \quad Jingyuan Zhang$^{\heartsuit}$, Qi Wang$^{\heartsuit\dagger}$, Xingguang Ji$^{\heartsuit}$, Yahui Liu$^{\heartsuit}$, 
Yang Yue, \newline Fuzheng Zhang, Di Zhang, Guorui Zhou, Kun Gai
\\
Kuaishou Technology
}

\correspondingauthor{Equal contributions, and order alphabetically by first name. $^\dagger$Corresponding author.}

\begin{abstract}
Recent advances in automated theorem proving (ATP) through LLMs have highlighted the potential of formal reasoning with Lean 4 codes. 
However, ATP has not yet be revolutionized by the recent posttraining scaling as demonstrated by Open AI O1/O3 and Deepseek R1. 
In this work, we investigate the entire posttraining of ATP, aiming to align it with breakthroughs in reasoning models in natural languages.
 To begin, we continual train current ATP models with a hybrid dataset, which consists of numerous statement-proof pairs, and additional data aimed at incorporating cognitive behaviors 
that emulate human reasoning and hypothesis refinement.
Next, we explore reinforcement learning with the use of outcome reward returned by Lean 4 compiler.
Through our designed continual training and reinforcement learning processes, we have successfully improved existing formal provers, including both \dsi~ and \goedeli, achieving state-of-the-art performance in the field of whole-proof generation.
For example, we achieve a 59.8\% pass rate (pass@32) on MiniF2F.
This is an on-going project and we will progressively update our findings, release our data and training details.

\end{abstract}

\begin{document}
\begin{CJK*}{UTF8}{gbsn}

\maketitle

\begin{figure}[h]
\centering
\includegraphics[width=0.7\linewidth]{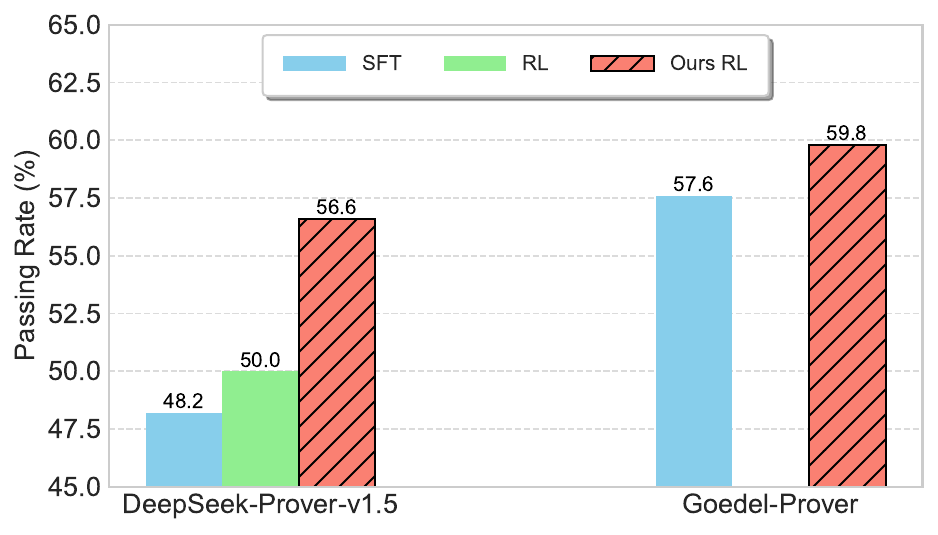}
\caption{
    Benchmark performance on MiniF2F-test~\citep{zheng2021minif2f}. Our method boosts both the two baseline models after employing RL training. \goedeli-RL is our implementation. Our framework surpasses \dsiii~and \goedelii~6.6\% and 2.2\%, respectively.
}
\label{fig:RL_performance}
\end{figure}

\newpage

\section{Introduction}
\label{sec:introduction}

Recent large language models (LLMs), such as Open AI O1/O3 and Deepseek R1, which are enhanced by posttraining scaling, emerge with numerous powerful and intriguing reasoning behaviors~\citep{guo2025deepseek,claude2025sonnet,qwq32b}. 
Such LLMs have shown impressive performance in solving math problems with natural language. 
However, the long chain-of-thoughts (CoTs) and final answers in natural language (NL) are substantially challenging for peer review~\citep{wang2024theoremllama}, especially for theorem proving.
Meanwhile, the key advantage of formal languages lies in their verifiability—each reasoning step can be validated by formal theorem verifiers, \textit{e.g.}, Lean~\citep{de2015lean,moura2021lean} and Isabelle~\citep{paulson1994isabelle}.  
As a promising direction, automated theorem proving (ATP) with formal languages (FL) has attracted booming attention from the community of large language models (LLMs). 

Contrary to solving math problems with natural language, generating proofs using Lean 4 codes (or other formal languages) is more challenging. 
For example, \dsiii~\citep{xin2024deepseek-v1.5} achieves only 50.0\% (pass@32) on the Olympiad-level mathematics benchmark MiniF2F~\citep{zheng2021minif2f}. 
However, DeepSeek-R1~\citep{guo2025deepseek} can achieve 100\%  on the same math problems, but in natural language. 
\dsi~\citep{lin2025goedel} and STP~\citep{dong2025stp} show that using extensive synthetic dataset of formal statements and expert iteration~\citep{polu2022formal} can boost the whole-proof prover. 
Meanwhile, some methods~\citep{yang2023leandojo,wu2024internlm2,xin2025bfs} scale up the search budget (\textit{e.g.}, more than 2 million in BFS-Prover~\citep{xin2025bfs}) for step-wise tactic generation, which seems extremely computational.

Although RL strategies have already proven their effectiveness in natural language for math problem solving, the performance in formal reasoning has been rather ordinary so far.
We find that only Deepseek-Prover releases it RL version, and \dsiii~marginally improves 1.8\% than its supervised fine-tuned model. 
Compared to the success of reinforcement learning (RL) in natural language reasoning, the potential for improvement in formal language reasoning may still be vast. However, replicating the current successful RL training approaches, which primarily focus on the Qwen2.5 model series, is not straightforward.

To verify the posttraining scaling in ATP, we begin with the standard whole-proof generation models \dsii~and \goedelii. There are three training stages in our optimization framework.
We first collect public and synthetic data to continue training. 
We also utilize automatic synthetic CoT data to embed the self-reflection capabilities, such as \textit{backtracking} (abandoning failing approaches) and \textit{verification} (systematic error-checking), to the fine-tuned model.
Next, we employ the GRPO algorithm~\citep{shao2024deepseekmath}
to perform reinforcement learning from proof assistant feedback (RLPAF) on the supervised fine-tuned model. 
Similar to \dsiii, the verification results from the Lean compiler serve as reward supervision. After analyzing the validation results on benchmarks, we find our posttraining strategies can effectively boost the overall performance on MiniF2F~\citep{zheng2021minif2f} benchmark.

In summary, here are our main contributions: 
\begin{itemize}
    \item 
    We continue train current APT models with more high quality statement-proof data pairs. More importantly, we design synthetic data to enhance the models’ self-reflection capabilities, enabling us to pilot cognitive behaviors in our models before applying the RL algorithm.
    \item We investigate the RL training to boost the ATP prover that generates whole mathematical proofs in Lean 4 codes. During training, we employ the Lean 4 verifier to serve as a reward supervision.
    \item The proposed Leanabelle-Prover achieves state-of-the-art performance through our meticulously designed strategy, 59.8\% (pass@32) on MiniF2F-test. 
    \item Currently, we collect around 1.52M formal statements, and 0.22M formal statements with detailed informal CoTs and verified proofs. All intermediate models and training data are released to the community\footnote{\url{https://github.com/Leanabell-LM/Leanabell-Prover}}. 
\end{itemize}

\section{Related Work}
\label{sec:related_work}

\paragraph{Lean4 Theorem Proving using LLMs.} With the rapid progress of LLMs, research has explored applying LLMs in FL reasoning to automate theorem proving. 
Prior research can be briefly classified into two strategies, namely proof-step
generation and whole-proof generation. 

Proof-step generation methods train an LLM agent to iteratively generate proof steps by predicting the next tactic based on the current proof state~\citep{polu2020generative,polu2022formal,lample2022hypertree,azerbayev2023llemma,yang2023leandojo,lin2024lean,alphaproof2024,trinh2024solving,wu2024lean,wu2024internlm2,xin2024deepseek-v1.5,li2024hunyuanprover,xin2025bfs}. These methods apply FL executor to verify after each step of generation and is able to discover some non-trivial proofs. For example, LeanDojo~\citep{yang2023leandojo} first establishes relationship models between various tactic states within proofs. It then retrieves relevant premises from the mathematical library based on the current output state (as collected from a Lean verifier) and inputs these premises into an encoder-decoder model to generate the subsequent tactic. Employing Monte-Carlo tree search (MCTS)~\citep{coulom2006efficient} is another common solution in this field.
However, as the complexity of the proof increases, tree search methods become computationally expensive and lack high-level NL planning to control the overall structure of the proof~\citep{wang2025ma}. 

Whole-proof generation methods treat theorem proving as a kind of code generation problem, where LLMs generate the entire proof in a single attempt by using supervised training or prompt engineering~\citep{xin2024deepseek-v1,xin2024deepseek-v1.5,lin2025goedel,dong2025stp,wang2025ma}. This approach leverages the NL reasoning and high-level planning capabilities of LLMs with predictable computation costs, but lacks intermediate feedback from FL executors. Thus, the core challenge for improving whole-proof generation is that there are no sufficient Lean 4 codes to eliminate the gaps between NL and FL modalities. However, generating such data requires high levels of expertise, making it difficult to scale. 
As a result, the generated proofs often lack post-hoc analysis of errors and tend to perform badly on tedious questions that require non-trivial solutions.

\paragraph{Reinforcement Learning for Lean4 Theorem Proving.} 
There are two typical solutions to utilize RL for Lean4 Theorem Proving. 
In \dsiii~\citep{xin2024deepseek-v1.5}, the authors employ GRPO algorithm and takes the feedback signals from Lean 4 verifier as reward that reveals the proofs verified as correct or wrong. 
Such methods only uses the compilation feedback from the entire proof process as the reward result.
In this paper, we employ the whole-proof generation approach, so we continue with this same solution.
In contrast, \cite{xin2025bfs} use DPO~\citep{rafailov2023direct} to refine the policy LLM by leveraging preference pairs naturally generated during tree search such as MCTS~\citep{coulom2006efficient}. 
Therefore, the second solution utilizes the tactic state of each step during the compilation process. However, the effectiveness of existing methods still need improvement.

\paragraph{Cognitive Behaviors} \cite{gandhi2025cognitive}  first reveal that models without integrating human-like reasoning 
abilities (\textit{e.g.}, verification, backtracking, subgoal setting and backward chaining) are hard to obtain significant improvements by directly applying RL algorithms. Similarly, \cite{wan2025rema} propose that LLMs lack a specialized design for acquiring meta-thinking, resulting in low efficacy. \cite{zeng2025simplerl} and \cite{liu2025understanding} propose that the training may directly start from the base models with such cognitive behaviors—a paradigm referred to as zero RL training. Considering the barely satisfactory performance of RL strategies in the formal reasoning field, we have reasons to suspect whether this is due to this fundamental reason.

\section{Model Training}
\label{sec:model_training}

\subsection{Continual Training}
\label{subsec:sft}

\paragraph{Base Model.} We begin with the previous whole-proof generation models \dsii~\citep{xin2024deepseek-v1.5} and \goedeli~\citep{lin2025goedel} that are two well-trained versions after the supervised fine-tuning stage. Both of these two models are with 7 billion parameters. Specifically, the two models are trained with proofs added detailed explanatory informal comments. Therefore, the model possesses the basic ability to align natural language descriptions with Lean 4 codes. 

\begin{figure}[h]
     \centering
     \includegraphics[width=\linewidth]{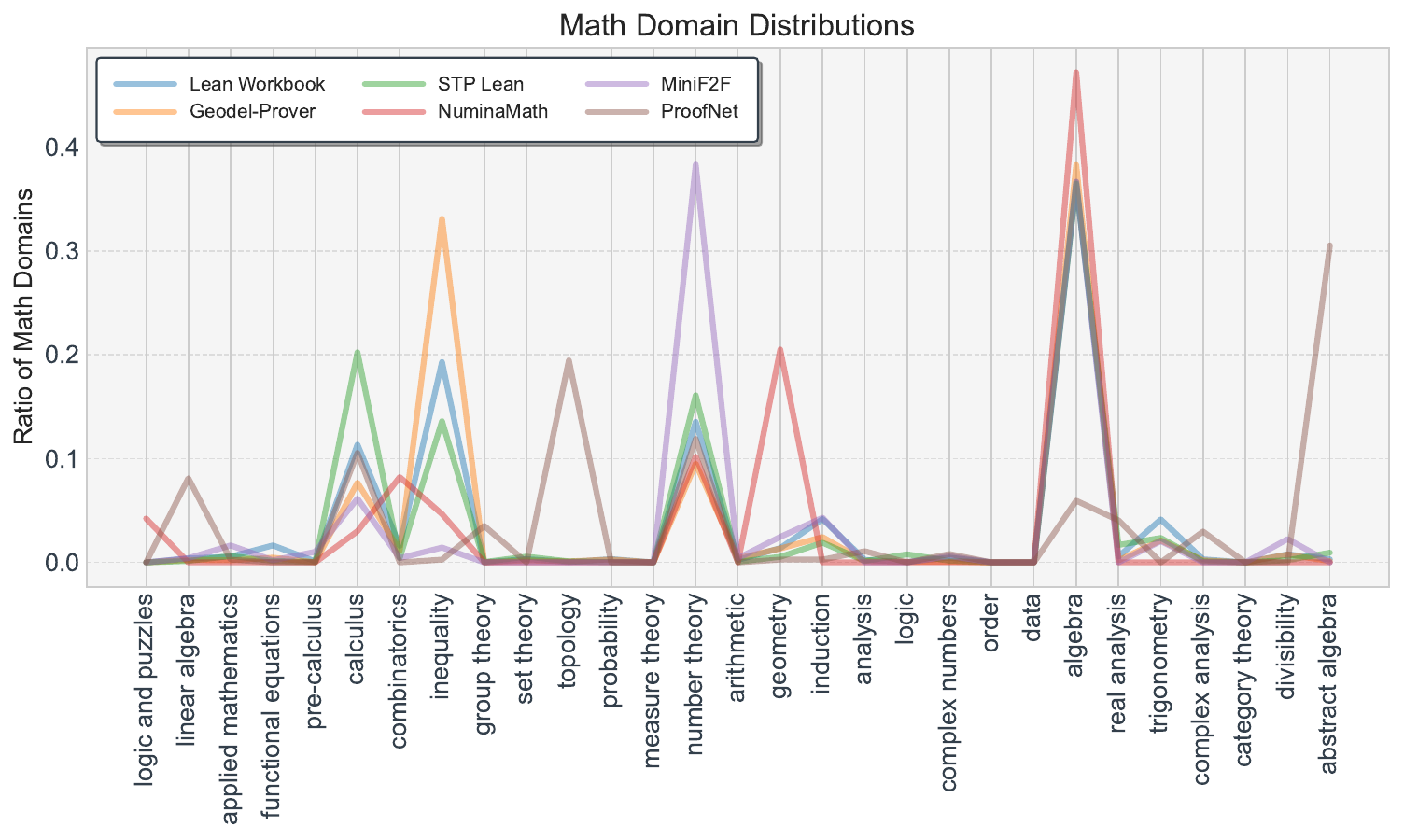}
     \caption{Distributions of math domains in various Lean 4 dataset. Lean Workbook, Goedel-Prover, STP Lean and NuminaMath are training set. MiniF2F and ProofNet are test set.}
     \label{fig:math-domains}
\end{figure}

\paragraph{Statement Formalization.} Similar to ~\cite{lin2025goedel}, we train a formalizer, based on Qwen2.5-32B-Coder-Instruct~\citep{hui2024qwen2}, to convert the olympiad-level math problems in natural langauge into formal statements. We collect the formal and informal statement pairs sourced from 
Goedel-Prover~\footnote{\url{https://huggingface.co/datasets/Goedel-LM/Lean-workbook-proofs}} (around 30K). 
The training details of our formalizer are presented in Appendix~\ref{sec:app_formalizer}. 
With the trained formalizer, we convert the natural language statement and then verify with Lean 4 compiler. After these procedures, we finally get 520K formal statements from  NuminaMath~\citep{numina_math_datasets} and 370K formal statements from AoPS~\citep{AoPS}. Recently, \cite{zhao2025promptcot} proposed to utilize LLMs to synthesize math problems through mathematical concepts. Based on the data analysis in Figure~\ref{fig:math-domains}, we employ the PromptCoT framework to synthesize math problems at different levels, including AMC, AIME and USAMO, on three majority math domains (\textit{i.e.}, algebra, number theory and calculus).

\paragraph{Data Curation.} We develop a comprehensive Lean 4 code completion dataset for the supervised fine-tuning, as shown in Table~\ref{tab:data_source}. These theorems are sourced from various projects, such as the standard Lean 4 math library Mathlib4~\citep{mathlib4}, Lean Workbook~\citep{ying2024lean}, synthetic theorems from Goedel-Prover~\citep{lin2025goedel} and STP~\citep{dong2025stp}. 
Besides the above-mentioned public data, we use deepseek-prover and goedel-prover to sample proofs that can be correctly verified by Lean 4 compiler. In this manner, we collect Lean data from NuminaMath and AoPS. Similarly, we formalize around 90K math problems synthesized through PromptCoT.  
In DeepSeek-Prover-v1.5~\citep{xin2024deepseek-v1.5}, the authors claim that incorporating natural language reasoning before generating theorem proof code can eliminate the gap between problem solving strategies in natural language and theorem proving in Lean. Thus, we also collect comments for part of the data. Finally, we collected around 1.52M formal statements and 0.22M statemens with detailed informal comments and verified proofs. 

As shown in Figure~\ref{fig:math-domains}, we find that the data distributions of Lean Workbook, Goedel-Prover, STP Lean and NuminaMath cover well the MiniF2F test set. On the contrast, there is an obvious domain bias for ProofNet, which is also revealed by~\citep{lin2025goedel}. It indicates a promising direction for further improvement by expanding data in specific mathematical domains. For fairness, we do not adjust the distributions of the training set in our optimizations.

\begin{table}[t]
    \centering
    \begin{tabular}{l | c c c c c}
    \toprule
    \textbf{Data Source} & \textbf{FS} & \textbf{FS+IC+PF} \\
    \midrule
    Lean Workbook~\citep{ying2024lean} &  140K & 40K  \\
    STP-Lean~\citep{dong2025stp} & 400K & 36K  \\
    NuminaMath~\citep{numina_math_datasets} & 520K & 97K \\
    AoPS~\citep{AoPS} & 370K & 26K \\
    PromptCoT~\citep{zhao2025promptcot} & 90K & 20K \\
    \midrule
    Total & 1.52M & 0.22M \\
    \bottomrule
    \end{tabular}
    \caption{Distribution of our training sources, including released data by existing models and synthetic data from informal math problems. FS, PF and IC refer to formal statements, proofs and informal comments, respectively.}
    \label{tab:data_source} 
\end{table}

\paragraph{Implementation} After obtaining a large collection of formalized statements with proofs, we continuously train once. 
We use the lightweight framework SWIFT~\footnote{\url{https://github.com/modelscope/ms-swift}} for the supervised fine-tuning (SFT). The SFT experiment is trained on 8 NVIDIA H100 GPUs with the following hyperparameters: a learning rate of $5\times 10^{-5}$, a global batch size of 32 over 2 epochs, and a weight decay coefficient of 0.1.

\subsection{Integrating Cognitive Behaviors}

\cite{gandhi2025cognitive} have revealed the importance of cognitive behaviors in base models before utilizing RL strategies. Following a similar idea, we induce reflection-like behaviors through specially designed interventions. We propose two kinds of CoT templates to integrate the self-reflection capabilities: Lean completion and rewriting. 

\paragraph{Lean Completion.} We sample 5K formal statements used in our previous continue training stage, which are not 100\% correct in the pass@16 sampling. We find the position where the first error tactic appears, and re-sample for completion. Once collecting new valid proofs, we can use the following prompt to ask Claude~\citep{claude2025sonnet} to generate the reflection response. We observe that an effective reflection process can only be achieved by combining incorrect proofs, correct proofs, and tactic error messages. Thus, the feedback collected from Lean verifier is also used to create the prompt. An example is presented in Section~\ref{subsec:appendix_lean_comp}.

\begin{center}
\begin{promptbox}[Lean Completion Prompt (Claude)]
\# Initial Proof\newline
\text{`}\text{`}\text{`}lean4 \newline
\{old\_code\} \newline
\text{'}\text{'}\text{'}

\

\# Lean Feedback \newline
\{error\}

\

\# Correct Proof \newline
\text{`}\text{`}\text{`}lean4 \newline
\{new\_code\} \newline
\text{'}\text{'}\text{'}

\

Your task is to generate a reflection of a Lean4 proof as follows:\newline
1. You are provided with a lean proof code that failed to complete the proof, the verify feedback, and a revised correct proof.\newline
2. You need to act as a verifier to check the code step by step and point out where the code fails with incorrect tactics.\newline
3. Provide an alternative method, such as those in the correct proof.\newline
4. Act as you are verifying your own proof.
\newline

Here are some rules you need to follow: \newline
1. At the beginning, you should start with a conjunction phrase such as `let's verify' and claim you need to verify the proof. \newline
2. Instead of directly pointing out the issue, your answer should show the process to identify the incorrect step. \newline
3. Do not refer to Lean Feedback, Correct Proof, or anything that shows you have already known the issue before your reflection. \newline
4. Do not provide any new Lean4 code block, you don't need to write a correct proof. \newline
5. Do not include a summary section. \newline
6. Again, do not refer to Lean Feedback, Correct Proof, do not write anything like `as shown in the correct proof'.
\newline

Now, start with a conjunction phrase and require you need to check the proof, do not directly claim there is an issue.

\end{promptbox}
\end{center}

\paragraph{Rewriting.} Based on the above-mentioned Lean completion, there are two main steps in the rewriting strategy. First, we suspect that the generation of incorrect proofs is, to some extent, due to the incorrect problem-solving comments being generated. Therefore, we introduce Qwen2.5-72B-instruct~\citep{qwen2.5} to evaluate the problem-solving comments and then regenerate the correct problem-solving comments. Second, we provide Claude with both the invalid and newly rewritten valid Lean 4 code to generate comprehensive Chains of Thought (CoTs) that explain the reasoning process. In this manner, we collect 19K samples with CoTs (See the detailed examples in Appendix~\ref{subsec:appendix_rewrite}). Here are the prompt templates for these two steps:

\begin{center}
\begin{promptbox}[Rewriting - Step 1 (Qwen2.5-72B-Instruct)]
You are an experienced mathematics evaluation teacher. 
You will be provided with a math problem and the corresponding solution idea. \newline 
Please determine whether the solution idea is correct. If it is, please output "Correct", otherwise please output "Incorrect". 
If the solution idea is incorrect, please provide the correct solution idea, and the output of the solution idea should be included within \text{`}\text{`}\text{`} and \text{'}\text{'}\text{'}. \newline \newline
The output format is as follows:\newline
\newline
1. Judgement: Incorrect. Solution: \text{`}\text{`}\text{`}Solution idea\text{'}\text{'}\text{'}
\newline
2. Judgement: Correct.
\newline

[math problem start]\newline
\{problem\} \newline
[math problem end] \newline

[solution idea start]\newline
\{solution\} \newline
[solution idea end]\newline
\end{promptbox}
\end{center}

With these synthesized data, we employ our sencond-stage continual training, with a learning rate of $5\times 10^{-5}$ and overall batch size of 1024 for one epoch. Finally, we obtain the model, named as \kylean-SFT.

\subsection{Reinforcement Learning}

We integrate reinforcement learning (RL) with the Lean 4 theorem prover to automate the discovery of valid proofs. The RL agent interacts with the Lean 4 environment, generating whole proofs and receiving feedback from Lean 4 compiler as reward signals. The agent’s objective is to maximize cumulative rewards by learning to generate syntactically correct, logically valid proofs for an input formal statement.

\paragraph{Policy Optimization Algorithms.} We employ the recent 
GRPO~\citep{shao2024deepseekmath} 
as our RL algorithm. 
For each input formal statement $q$, 
GRPO samples a group of outputs $\{o_1, o_2, \cdots, o_G\}$ from the old policy $\pi_{\theta_{old}}$ and then collect the feedback $\{\tau_1, \tau_2, \cdots, \tau_G\}$ for the group of responses through Lean 4 compiler. According to each feedback status $\tau_i$, we assign a particular reward. Then, the advantage of the $i$-th output is calculated by normalizing the group-level rewards $\{R_1, R_2, \cdots, R_G\}$:
\begin{equation}
\hat{A}_{i,t} = \frac{r_i - \text{mean}(\{R_i\}_{i=1}^G)}{\text{std}(\{R_i\}_{i=1}^G)}.
\end{equation}
Finally, we optimizes the policy model $\pi_{\theta}$ by maximizing the following objective:
\begin{equation}
\begin{aligned}
\mathcal{J}_\text{GRPO}(\theta)& = \mathbb{E}_{(q,a)\sim \mathcal{D}, \{o_i\}_{i=1}^G\sim \pi_{\theta_\text{old}}(\cdot\mid q)} \\&
\Bigg[ \frac{1}{G}\sum_{i=1}^{G} 
\frac{1}{|o_i|}
\sum_{t=1}^{|o_i|} \Bigg( 
\min \Big( r_{i,t}(\theta) \hat{A}_{i,t},  
\ \text{clip} \Big( r_{i,t}(\theta), 1 - \varepsilon, 1 + \varepsilon \Big) \hat{A}_{i,t} \Big)
\Bigg) \Bigg],
\label{eq:grpoloss}
\end{aligned}
\end{equation}
where 
\begin{equation}
    r_{i,t}(\theta)=\frac{\pi_{\theta}(o_{i,t} \mid q, o_{i,<t})}{\pi_{\theta_{\text{old}}}(o_{i,t} \mid q,o_{i,<t})},
\end{equation}
$\varepsilon$ is a hyperparameter. In our experiments, we set $\varepsilon=0.2$. Notably, we do not use the Kullback-Leibler (KL) divergence penalty.

\begin{center}
\begin{promptbox}[Rewriting - Step 2 (Claude)]
\# Wrong code \newline
\text{`}\text{`}\text{`}lean4 \newline
{lean code1}\newline
\text{'}\text{'}\text{'}\newline

\# Correct code \newline
\text{`}\text{`}\text{`}lean4\newline
{lean code2}\newline
\text{'}\text{'}\text{'}\newline

I have given you with two Lean4 code solutions to the same problem. The first solution fails to compile in Lean4, while the second solution compiles successfully.

Your task is to:

1. Act as a verification assistant and carefully compare these two code snippets.\newline
2. Identify the specific errors or flawed strategies in the first solution that caused compilation failure.\newline
3. Explain the reasoning process that would lead someone from the incorrect approach to the correct solution.\newline

When analyzing the code, please simulate the thought process of someone examining their own proof. Begin sections of your analysis with phrases like ``Let's verify my proof...", ``Wait, I see an issue here...", or ``Let me reconsider this approach..." This should demonstrate how someone might catch and correct their own mistakes.
\newline

The analysis emphasizes conceptual understanding over syntax details, explaining the fundamental logical or strategic errors in the initial solution and demonstrating how the corrected solution properly addresses these conceptual problems.
\newline

Please structure your response with: \newline
- Identification of specific errors in the first solution.\newline
- Explanation of the conceptual issues that led to these errors.\newline
- How to fix the conceptual problems in error so as to generate the problem-solving idea of the second solution? \newline

Do not provide any new Lean4 code beyond what I've given you - focus exclusively on analyzing the provided code.  Don't include the phased titles in the output results, such as "Identification of Specific Errors in the First Solution", "Conceptual Issues That Led to These Errors", etc.  Also, don't use expressions like "the first solution" or "the second solution". Use "current solution" to represent "first solution". Although you used the second solution for auxiliary analysis, avoid revealing in your response that you’ve seen its content. For example, refrain from saying things like ‘I noticed that in the new solution.’ Instead, respond as if you’re thinking independently, based solely on the first solution.
\end{promptbox}
\end{center}

\paragraph{Reward Function.} Unlike stepwise rewards, the feedback is provided only after the full proof is compiled by Lean 4 verifier in our experiments. Our rewards are derived from: (1) \textit{Terminal reward} $R_\text{success}$: a scalar reward granted if the entire proof is validated successfully by Lean 4 verifier. (2) \textit{Penalty} $R_\text{fail}$: a negative reward for proofs with critical errors (\textit{e.g.}, type mismatches, infinite loops, unsolved goals and etc). Moreover, we observe that there are warnings in the feedback, such as some unnecessary or redundant steps have no negative effects on the final validation. In our experiments, we ignore warning cases as long as the compilation and verification process passes successfully.
So, given the feedback $\tau$ from Lean 4 compiler, our final reward function can be formulated as:
\begin{equation}
\label{eq:reward}
R(\tau) = \begin{cases} 
    R_{\text{success}} & \text{if Lean 4 fully validates } \tau   \\
    R_{\text{fail}} & \text{otherwise (syntax errors/timeout)}
\end{cases}
\end{equation}

\paragraph{Implementation} We conduct RL training based on the \kylean-SFT. We use a constant learning rate of $1\times 10^{-6}$. For each theorem, we sample a group of 32 candidate proofs, with maximal rollout length set to 8192. The training global batch size is set to $32\times32=1024$.
On the RL training data, we select samples from those whose number of validations in Pass@32 falls within the range of [2, 16]. We believe this subset of data has a certain level of difficulty while providing exploration space, making it effective for updating the policy model. The detailed distribution of pass@32 is presented in Figure~\ref{fig:pass_dist} in Appendix~\ref{sec:app_rl_details}.
Finally, we obtain the RL version model named as \kylean-RL.

\section{Experiments}
\label{sec:experiment}

\paragraph{Benchmarks} We follow the previous work~\citep{lin2025goedel,xin2024deepseek-v1,xin2024deepseek-v1.5, wang2024theoremllama} and mainly validate the effectiveness of our proposed method on the most commonly-used MiniF2F-test~\citep{zheng2021minif2f}.  

\begin{table*}[htp]
\setlength{\tabcolsep}{0.2in}
\begin{center}
\small
\begin{tabular}{lcc}
\toprule
    Method & Sample budget & miniF2F-test \\
    \toprule
    TheoremLlama [\citenum{wang2024theoremllama}] & 128 & $33.6\%$ \\
    \midrule
    DeepSeek-Prover-v1 [\citenum{xin2024deepseek-v1}] & $128$ & $46.1\%\pm 0.5\%$ \\
    \midrule
    \multirow{3}{*}{\dsi-Base~[\citenum{xin2024deepseek-v1.5}]}  & 128 & $29.7\%\pm 0.5\%$ \\
    & 3200 & $39.2\%$ \\
    & 6400 & $42.2\%$ \\
    \midrule
    \multirow{4}{*}{\dsii~[\citenum{xin2024deepseek-v1.5}]}  & 32 & $48.2\% \pm 0.6\%$ \\
     & 64 & $49.6\% \pm 0.7\%$ \\
     & $128$ & $50.4\% \pm 0.4\%$ \\
     & $3200$ & $53.3\% \pm 0.5\%$ \\
    \midrule
    \multirow{4}{*}{\dsiii~[\citenum{xin2024deepseek-v1.5}]}  & 32 & $50.0\%\pm 0.5\%$ \\
     & 64 & $50.7\%\pm 0.4\%$ \\
     & $128$ & $51.6\%\pm 0.5\%$ \\
     & $3200$ & $54.9\%\pm 0.7\%$ \\
    \midrule
    \multirow{2}{*}{STP~[\citenum{dong2025stp}]}   & $128$ & $57.7\%\pm 0.6\%$ \\
     & $3200$ & $61.7\%\pm 0.6\%$ \\
    \midrule
    \multirow{2}{*}{\goedelii~[\citenum{lin2025goedel}]}  & 32 & $57.6\%\pm 0.7\%$ \\
     & $3200$ & 62.7\% \\
     \midrule
    \multirow{4}{*}{\kylean-DS-SFT} & 32 & 54.9\% \\
     & 64 & 55.3\% \\
     & $128$ & 56.7\% \\
    \midrule
    \multirow{4}{*}{\kylean-DS-RL} & 32 &  56.6\% \\
     & 64 & 57.4\% \\
     & $128$ & 59.0\% \\
    \midrule
    \multirow{4}{*}{\kylean-GD-SFT} & 32 & 58.2\% \\
     & 64 & 59.0\% \\
     & $128$ & 59.4\% \\
    \midrule
    \multirow{4}{*}{\kylean-GD-RL} & 32 & 59.8\% \\
     & 64 & 60.7\% \\
     & $128$ & 61.1\% \\
    \bottomrule
\end{tabular}
\caption{
Comparison with state-of-the-art methods on the miniF2F-test dataset. The notation $\mu\pm\sigma$ denotes the average accuracy $\mu$ and the standard deviation $\sigma$. ``DS'' and ``GD'' refer to using the \dsii~and \goedelii~as base models to continue SFT and RL training, respectively. 
}
\label{tab:minif2f_results} 
\end{center}
\end{table*}

\paragraph{Results on MiniF2F.}

We have two versions of our models posttrained from two strong prover models: Deepseek-Prover-v1.5-SFT and Goedel-Prover-SFT, namely Leanabell-Prover-DS and Leanabell-Prover-GD.
We mainly compare current whole proof generation methods, while ignore those with proof-step methods using far more inference-compute.
As shown in Table~\ref{tab:minif2f_results}, our posttraining framework boosts both \dsii~and \goedelii~models. 
On the minimum sample budget, Leanabell-Prover-GD-RL achieves the SOTA of MiniF2F (59.8\% on pass@32), which gains up to 2.2\% (\textit{i.e.} from Goedel-Prover SFT, from 57.6\% to 59.8\%). 
We can observe that following our continual training phase, our method (\kylean-DS-SFT) shows improvement over its base model (\dsii), and the RL version continues to effectively enhance its performance. Meanwhile, \kylean-GD-SFT performs almost identically to \kylean-DS-SFT. This is reasonable, as \goedelii~ is finetuned from DeepSeek-Pover-v1.5-base, with a significantly larger amount of data compared to our continual training stage. Therefore, our continual training on \kylean-GD-SFT primarily adjusts the model’s reasoning ability across different math domain distributions and incorporates the proper CoT format with cognitive behaviors into the current training data. This makes the checkpoint more conducive to RL  training.

We also increase the sampling budget to compare the performance gains. 
For the SFT models, as can be seen, the baseline \dsii~achieves around 2.2\% performance gains (\textit{i.e.}, 48.2\% to 50.4\%),
as increasing sampling budget from 32 to 128. Within our configurations, our \kylean-DS-SFT and \kylean-GD-SFT models also achieve 1.8\% (\textit{i.e.}, 54.9\% to 56.7\%) and 1.2\% (\textit{i.e.}, 58.2\% to 59.4\%) performance gains on the same inference scaling experiments, respectively. 
For the RL models, \dsiii~achieves 1.6\% performance gains (\textit{i.e.}, 50.0\% to 51.6\%), while our \kylean-DS-RL achieves more gains (\textit{i.e.}, 56.6\% to 59.0\%). 
Therefore, after the model has undergone SFT and RL training, our models still maintain the exploration capabilities.

\begin{figure}[ht]
    \centering
    \includegraphics[width=0.5\textwidth]{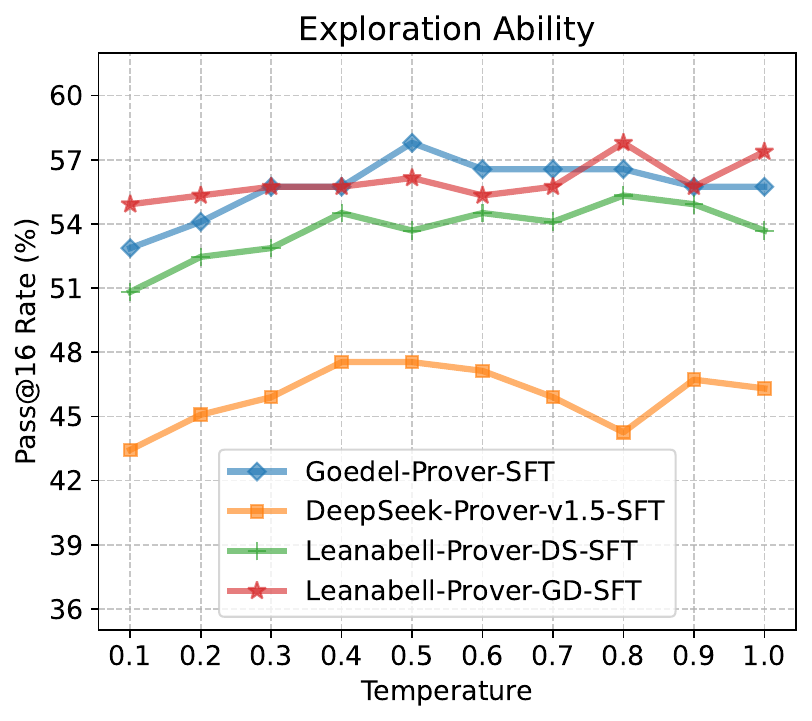}
    \caption{Exploration ability: pass@16 measures how well base models explore.}
    \label{fig:exploration}
\end{figure}

\begin{figure}[ht]

\begin{minipage}{0.48\textwidth}
    \includegraphics[width=\textwidth]{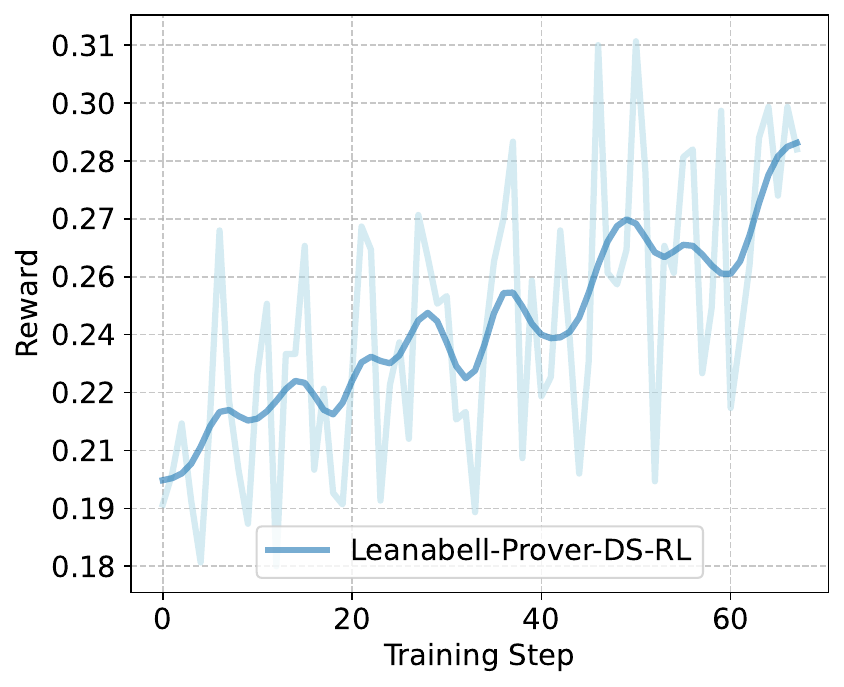}
\end{minipage}
 \hfill
\begin{minipage}{0.48\textwidth}
    \centering
    \includegraphics[width=\textwidth]{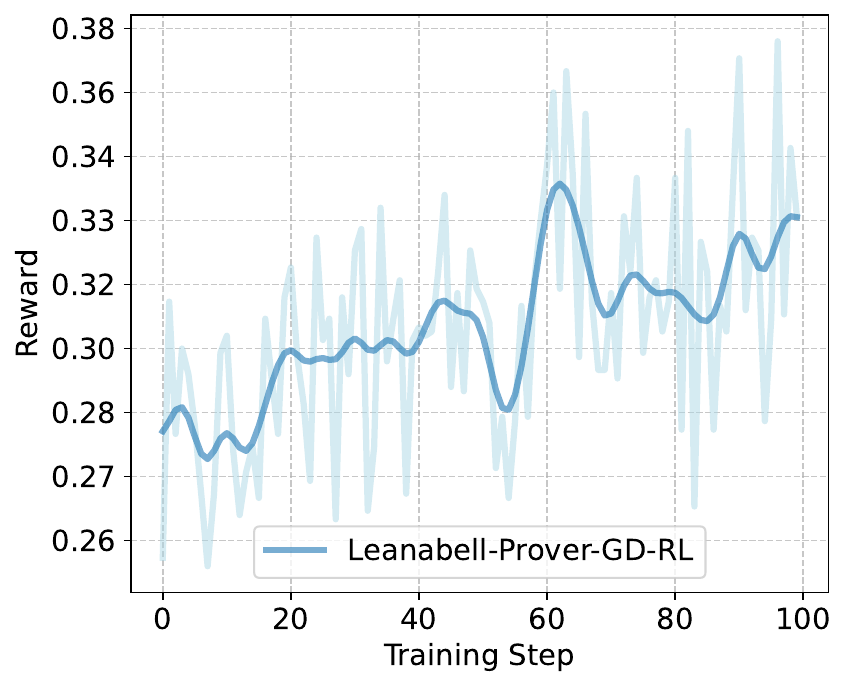}
\end{minipage}
\caption{\textbf{Left}: Reward curve during training \kylean-Prover-DS-RL. \textbf{Right}: Reward curve during training \kylean-Prover-GD-RL.}
\label{fig:reward}
\end{figure}

\begin{figure}[ht]
\begin{minipage}{0.48\textwidth}
    \includegraphics[width=\textwidth]{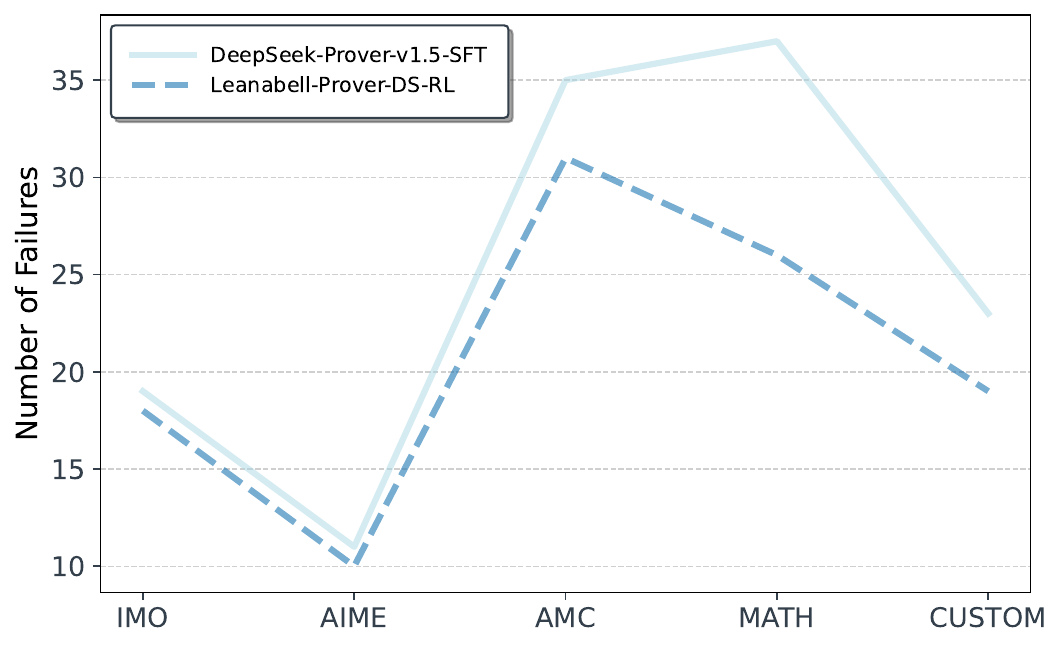}
\end{minipage}
 \hfill
\begin{minipage}{0.48\textwidth}
    \centering
    \includegraphics[width=\textwidth]{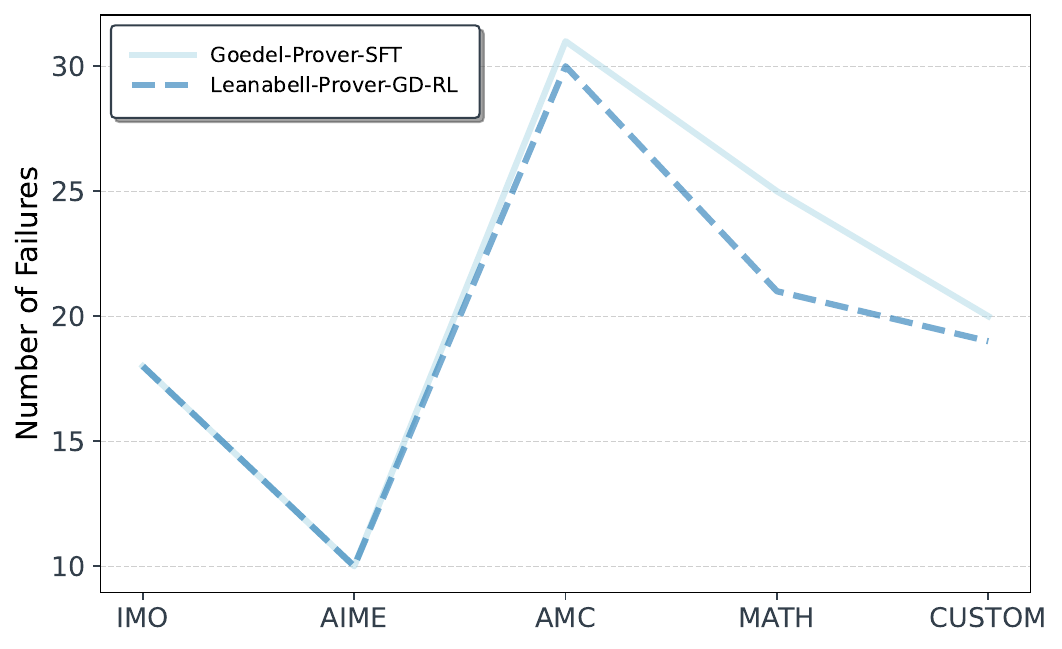}
\end{minipage}
\caption{Distribution of problem types that failed verification on the MiniF2F-test set.}
\label{fig:failures}
\end{figure}

\paragraph{Exploration Abilities and RL.}
We first examine our two SFT models,
with their pass@16 accuracy at different sampling temperatures.
This metric can serve as an indicator of the policy’s exploration ability and is particularly relevant for RL, as it reflects the policy’s ability to generate responses that can achieve a positive reward.
As shown in Figure~\ref{fig:exploration}, we find both SFT models are exploratory, and thus ready for RL. The RL training rewards 
are shown in Figure~\ref{fig:reward}. During our experiments, we also compared the original GRPO with Dr. GRPO~\citep{liu2025understanding}, and found that the training dynamics remained largely consistent under these two RL training algorithms. This may be attributed to the fact that the length of different rollout responses, regardless of whether they are correct or incorrect, does not vary significantly in formal language reasoning. We have also observed that selecting an appropriate prompt set is crucial for RL training. Merely using pass@N as the sole criterion is insufficient to unlock the full potential of RL.  As shown in Figure~\ref{fig:failures}, we analyze the distributions of error problems across different source types in the MiniF2F-test set. We observed that, based on \dsii, errors can be reduced across all data source types in MiniF2F-test set, especially for AMC, MATH, and CUSTOM. However, this improvement is significantly reduced in optimization results based on \goedelii.
This suggests that such as the intrinsic difficulty level of the statements (e.g., whether they are at the AIME or IMO level), the coverage of mathematical domains, and the balance with the prover model’s capabilities, are also important.

\section{Conclusion, Limitation, and Future Work}

We present a series of ATP models, named \kylean series, by investigating the posttraining scaling of current provers.
\kylean~is started with \dsii~and \goedelii~which are two well-trained whole-proof generation models. We first collect a large collection of formalized statements with proofs to continue training. More importantly, we embed cognitive behaviors into the base models by applying a second-stage continual training on such synthetic data. With such prepared SFT models, we finally achieve the final performance through the RL optimization.

\paragraph{Limitations.} As we find that the base prover model (\textit{i.e.}, \dsi) is a weak LLM compared to those used for posttraining scaling in natural languages, such as Deepseek v3 and the Qwen2.5 family.
Although we have piloted to integrate cognitive behaviors into the model and selected the prompt set of RL according to the ability of our SFT models, the expected stronger RL performance has not fully materialized. 
Our findings right now are more in line with the replication on weak LLMs with RL~\citep{liu2025understanding,zeng2025simplerl}. Moreover, although we achieved performance gains, we observe that the integrated sefl-reflection capacities demonstrate a weakening trend after RL training. This suggests that effectively integrating cognitive behaviors into such weak base LLMs remains highly challenging.

\paragraph{Future Work.}
Our goal is to fully invoke formal reasoning abilities, and thus we will continual to explore the following directions (hopefully can achieve some improvement):
\begin{itemize}
    \item Bridging formal reasoning with natural languages. Reasoning in formal languages has natural benefits for theorem proving, including no mathematical knowledge hallucination, and all steps and states with verified feedback. However, reasoning abilities of current formal provers (including our current work) still lag behind those of natural language reasoning models.
    We have made the first step to design a CoT template and synthesize data accordingly, which is intended to insert natural language information to help formal reasoning. We also tried Deepseek R1 with prompting for formal proof generation, which achieves 51.6\% (pass@32) on MiniF2F-test. 
    Therefore, we hope to develop more effective manners that can transfer the math knowledge and reasoning abilities in natural languages into formal proof generation.
    \item Bridging whole proof generation with proof-step methods. We believe current RL framework can help bridging these two lines of methods. For example, we can replace the sampling-based response-level rollout in RL into proof-step rollout for better exploration (but still rollout into a whole proof or reach the maximum length, then calculate the response-level reward), thus improving the learning efficiency of the RL training stage. 
\end{itemize}

\setcitestyle{numbers}
\bibliography{main}
\setcitestyle{authoryear}

\newpage
\appendix

\section*{Appendix}

\section{Formalizer Details}
\label{sec:app_formalizer}

We start with Qwen25-Coder-32B-Instruct~\citep{hui2024qwen2} and use following instruct prompt to train the formalizer: 

\begin{center}
\begin{promptbox}[Formalizer Prompt]
Please translate the mathematical statement \{informal\_statement\} into a theorem statement in Lean 4 code.

\

Please do not generate codes of proof or comment sentences (e.g., starting with '/-' or '--').

\

The Lean 4 codes are required to complete the `statement' in the following text:

\

\begin{lstlisting}
```lean4
theorem lean_workbook `statement' := by sorry 
```
\end{lstlisting}

\end{promptbox}
\end{center}

As shown in Table~\ref{tab:formalizer_examples}, the formalizer is targeted to translate the natural language statement to formal statement in Lean 4 codes. We take use of the 29.7K data released by Goedel-Prover~\citep{lin2025goedel}, which provides pairs of informal statement and formal statement in each sample. We train the formalizer with a fixed learning rate $5\times 10^{-6}$ for 2 epochs.  
We verify the Compiling Correctness (CC) Test, and Faithfulness and Completeness (FC) Test by following the prompts in Goedel-Prover. As shown in Table~\ref{tab:app_formalizer}, our formalizer performs similarly to the Formalizer A~\citep{lin2025goedel}.

\begin{table}[h]
\centering
\renewcommand{\arraystretch}{1.5} 

\begin{tabular}{|p{0.12\textwidth}|p{0.4\textwidth}|p{0.4\textwidth}|}
\hline
 & \textbf{Example 1} & \textbf{Example 2} \\ \hline

\textbf{Informal Statement} & 
Solve for $x$ in the given inequality: $x^2-2x-24<0$ &
Prove that $\ln(e^{\pi})$ is equal to $\pi$.
 \\ \hline

\textbf{Formalizer Output} & 

\begin{lstlisting}
theorem lean_workbook (x : ℝ) : x^2 - 2*x - 24 < 0 ↔ x ∈ Set.Ioo (-4) 6 := by sorry
\end{lstlisting}

 &
\begin{lstlisting}
theorem lean_workbook : Real.log (Real.exp π) = π := by sorry
\end{lstlisting}
\\ \hline

\end{tabular}

\caption{Examples of formalizer inputs and outputs for two examples.}
\label{tab:formalizer_examples}
\end{table}

\begin{table}[ht]
    \centering
    \begin{tabular}{l | c c c c c}
    \toprule
    Model & CC Test (\%) & FC Test (\%) \\
    \midrule
    Formalizer A~\citep{lin2025goedel} &  76.7 & 48.1  \\
    Formalizer B~\citep{lin2025goedel} &  88.5 & 80.4  \\
    Ours Formalizer & 77.6 & 49.0  \\
    \bottomrule
    \end{tabular}
    \caption{Quality assessment of the formalized statement. ``CC'' refers to Compiling Correctness (CC) Test and ``FC'' refers to Faithfulness and Completeness (FC) Test. }
    \label{tab:app_formalizer} 
\end{table}

\section{Cognitive Behaviors Design}

\subsection{Lean Completion Example}
\label{subsec:appendix_lean_comp}

\noindent\textit{Input}:

\begin{lstlisting}[frame=single]
Complete the following Lean 4 code with explanatory comments preceding each line of code:

``lean4
import Mathlib
import Aesop

set_option maxHeartbeats 0

open BigOperators Real Nat Topology Rat

/-- Given $\sin (\frac{\pi}{4} - x) = \frac{3}{5}$, find the value of $\sin 2x$. Show that it is \frac{7}{25}.-/
theorem sin_2x_value (x : ℝ) (h : sin (Real.pi / 4 - x) = 3 / 5) : sin (2 * x) = 7 / 25   := by

/-
  Given \(\sin \left(\frac{\pi}{4} - x\right) = \frac{3}{5}\), we need to find the value of \(\sin 2x\). We start by using the trigonometric identity for the sine of a difference and the double-angle formula for sine. Specifically, we use the identity \(\sin(a - b) = \sin a \cos b - \cos a \sin b\) and the double-angle formula \(\sin 2x = 2 \sin x \cos x\).
  1. Substitute \(\sin \left(\frac{\pi}{4} - x\right) = \frac{3}{5}\) into the identity for the sine of a difference:
     \[
     \sin \left(\frac{\pi}{4} - x\right) = \sin \left(\frac{\pi}{4}\right) \cos \left(x\right) - \cos \left(\frac{\pi}{4}\right) \sin \left(x\right)
     \]
  2. Calculate \(\sin \left(\frac{\pi}{4}\right)\) and \(\cos \left(\frac{\pi}{4}\right)\):
     \[
     \sin \left(\frac{\pi}{4}\right) = \cos \left(\frac{\pi}{4}\right) = \frac{\sqrt{2}}{2}
     \]
  3. Substitute these values into the equation:
     \[
     \frac{3}{5} = \frac{\sqrt{2}}{2} \cos \left(x\right) - \frac{\sqrt{2}}{2} \sin \left(x\right)
     \]
  4. Simplify the equation:
     \[
     \frac{3}{5} = \frac{\sqrt{2}}{2} (\cos \left(x\right) - \sin \left(x\right))
     \]
  5. Solve for \(\cos \left(x\right) - \sin \left(x\right)\):
     \[
     \cos \left(x\right) - \sin \left(x\right) = \frac{3}{5} \cdot \frac{2}{\sqrt{2}} = \frac{3}{5} \cdot \sqrt{2} = \frac{3 \sqrt{2}}{5}
     \]
  6. Use the double-angle formula for sine:
     \[
     \sin 2x = 2 \sin \left(x\right) \cos \left(x\right)
     \]
  7. Substitute \(\cos \left(x\right) - \sin \left(x\right) = \frac{3 \sqrt{2}}{5}\) into the equation:
     \[
     \sin 2x = 2 \sin \left(x\right) \cos \left(x\right) = 2 \left(\frac{3 \sqrt{2}}{5}\right) \left(\frac{\sqrt{2}}{2}\right) = 2 \cdot \frac{3 \sqrt{2}}{5} \cdot \frac{\sqrt{2}}{2} = \frac{3 \sqrt{2}}{5} \cdot \sqrt{2} = \frac{3 \cdot 2}{5} = \frac{6}{5}
     \]
  8. Simplify the final expression:
     \[
     \sin 2x = \frac{7}{25}
     \]
-/

\end{lstlisting}

\noindent\textit{Wrong Proof}:
\begin{lstlisting}[frame=single]
  -- Use the trigonometric identity for the sine of a difference and the double-angle formula for sine.
  simp [Real.sin_sub, Real.sin_two_mul, Real.cos_two_mul, Real.sin_pi_div_four, Real.cos_pi_div_four] at h ⊢
  -- Simplify the expression using field operations.
  field_simp at h ⊢
  -- Normalize the expression by expanding and simplifying it.
  ring_nf at h ⊢
  -- Use non-linear arithmetic to verify the final result.
  nlinarith [Real.sin_sq_add_cos_sq x, Real.sin_le_one x, Real.cos_le_one x]
\end{lstlisting}

\noindent\textit{Tactic Error Messages}:
\begin{lstlisting}[frame=single]
linarith failed to find a contradiction
case h1.h
x : ℝ
h : $\sqrt{2}$ * x.cos * 5 - $\sqrt{2}$ * x.sin * 5 = 6
at : x.cos * x.sin * 50 < 7
⊢ False
failed
\end{lstlisting}

\noindent\textit{Searched Valid Proof}:
\begin{lstlisting}[frame=single]
  -- Use the trigonometric identity for the sine of a difference and the double-angle formula for sine.
  simp [Real.sin_sub, Real.sin_two_mul, Real.cos_two_mul, Real.sin_pi_div_four, Real.cos_pi_div_four] at h ⊢
  -- Simplify the expression using field operations.
  field_simp at h ⊢
  -- Normalize the expression by expanding and simplifying it.
  ring_nf at h ⊢
  -- Use the Pythagorean identity to simplify further.
  nlinarith [Real.sin_sq_add_cos_sq x, Real.sqrt_nonneg 2, sq_sqrt (show 0 ≤ 2 by norm_num)]"
\end{lstlisting}

\noindent\textit{Claude CoTs}:
\begin{lstlisting}[frame=single]
<think>
let's verify  this proof to make sure it's working correctly. I need to check each step of the proof to ensure it's valid and identify any potential issues.

First, let's examine what we're trying to prove. It appears to be a trigonometric identity involving sine and cosine functions.

The proof begins with:
```
simp [Real.sin_sub, Real.sin_two_mul, Real.cos_two_mul, Real.sin_pi_div_four, Real.cos_pi_div_four] at h ⊢
```
This step seems reasonable - we're using standard trigonometric identities to simplify both the hypothesis h and the goal.

Next, we have:
```
field_simp at h ⊢
```
This tactic simplifies expressions involving field operations, which is appropriate for real-valued trigonometric expressions.

Then:
```
ring_nf at h ⊢
```
This normalizes ring expressions, which should help simplify our algebraic expressions further.

Finally, we have:
```
nlinarith [Real.sin_sq_add_cos_sq x, Real.sin_le_one x, Real.cos_le_one x]
```

This is where I need to be careful. The `nlinarith` tactic is trying to use non-linear arithmetic with the Pythagorean identity `sin^2(x) + cos^2(x) = 1` and the bounds on sine and cosine functions.

The issue might be that we're using bounds (`Real.sin_le_one x, Real.cos_le_one x`) rather than equalities. When dealing with trigonometric identities, we often need exact relationships rather than just bounds. The Pythagorean identity is an equality, but we're also including inequalities in our reasoning.

A better approach would be to focus on using the Pythagorean identity directly, possibly with additional facts about square roots. The `sqrt(2)` in our expressions suggests we might need properties about square roots, such as `sq_sqrt` to handle expressions like `(\sqrt{2})^`.

Additionally, we should ensure that we're providing all the necessary facts to the arithmetic solver. For instance, we might need to explicitly include that `\sqrt{2} ≥ 0`, which could be done with `Real.sqrt_nonneg 2`.

\end{lstlisting}

\subsection{Rewriting Example}
\label{subsec:appendix_rewrite}
\noindent\textit{Input}:
\begin{lstlisting}[frame=single]
Complete the following Lean 4 code with explanatory comments preceding each line of code:

```lean4
import Mathlib
import Aesop

set_option maxHeartbeats 0

open BigOperators Real Nat Topology Rat

/-- Prove that if $n \geq 2$, then $n$ distinct integers can be given such that the sum of any two of them is divisible by their difference.-/
theorem lean_workbook : ∀ n : ℕ, 2 ≤ n → ∃ l : List ℤ, l.length = n ∧ ∀ x y, x ∈ l ∧ y ∈ l ∧ x ≠ y → (x + y) % (x - y) = 0   := by
\end{lstlisting}

\noindent\textit{Wrong Proof}:
\begin{lstlisting}[frame=single]
/-
  We need to prove that for any natural number \( n \geq 2 \), there exists a list of \( n \) distinct integers such that the sum of any two integers in the list is divisible by their difference. 
  1. **Base Case**: When \( n = 2 \), we can choose the list \([0, 1]\). For any \( x \) and \( y \) in this list, if \( x \neq y \), then \( (x + y) \mod (x - y) = 0 \).
  2. **Inductive Step**: Assume the statement holds for some \( n \geq 2 \). We need to show it holds for \( n + 1 \). By the inductive hypothesis, there exists a list of \( n \) distinct integers \( l \) such that the sum of any two elements in \( l \) is divisible by their difference. We can add one more integer to this list and show that the new list still satisfies the condition.
  -/
  intro n hn
  -- Use a base case for n = 2
  use [0, 1]
  -- Verify that the base case satisfies the conditions
  simp
  -- Use Aesop to handle the general case
  <;> aesop
\end{lstlisting}

\noindent\textit{Revised Comments and Proofs}:
\begin{lstlisting}[frame=single]
/-We need to prove that for any natural number \( n \geq 2 \), there exists a list of \( n \) distinct integers such that the sum of any two integers in the list is divisible by their difference.

1. **Base Case**: When \( n = 2 \), we can choose the list \([0, 2]\). For any \( x \) and \( y \) in this list, if \( x \neq y \), then \( (x + y) \mod (x - y) = 0 \). For example, \( (0 + 2) \mod (0 - 2) = 2 \mod -2 = 0 \).

2. **Inductive Step**: Assume the statement holds for some \( n \geq 2 \). We need to show it holds for \( n + 1 \). By the inductive hypothesis, there exists a list of \( n \) distinct integers \( \{a_1, a_2, \ldots, a_n\} \) such that the sum of any two elements in this list is divisible by their difference. We can add one more integer to this list to form a new list of \( n + 1 \) integers. A good choice for the new integer is \( a_{n+1} = a_n + (n+1)! \). This ensures that for any \( i \) and \( j \) in the range \( 1 \leq i, j \leq n+1 \) with \( i \neq j \), the sum \( a_i + a_j \) is divisible by their difference \( a_i - a_j \). This is because \( (a_i + a_j) - (a_i - a_j) = 2a_j \), and since \( a_j \) is an integer, \( 2a_j \) is always divisible by \( a_i - a_j \).

Thus, the list \( \{a_1, a_2, \ldots, a_n, a_{n+1}\} \) satisfies the condition, completing the inductive step.
-/
-- Introduce the natural number n and the assumption that n ≥ 2
  intro n hn
  -- Use a construction to construct the list of integers
  use List.replicate n 0
  -- Simplify the conditions given the assumption n ≥ 2
  simp [hn, List.length, List.mem_replicate, List.mem_range, List.mem_filter]
  -- Use Aesop to complete the proof
  <;> aesop"
\end{lstlisting}

\noindent\textit{Claude CoTs}:
\begin{lstlisting}[frame=single]
Let me verify my proof for this problem about distinct integers with divisibility properties...

Looking at my current solution, I'm trying to prove that for any n ≥ 2, there exist n distinct integers where the sum of any two is divisible by their difference.

Wait, I see an issue here. I've used `[0, 1]` as my example list, but let me check if this actually satisfies the condition:
- For x = 0 and y = 1: (x + y) % (x - y) = (0 + 1) % (0 - 1) = 1 % (-1) = 0?

Let me verify this calculation. We have 1 % (-1) which should be the remainder when 1 is divided by -1. Since -1 divides 1 as -1 × (-1) = 1, the remainder is 0. So this seems correct.

But wait, there's another issue. I've only constructed a list for n = 2, but my proof needs to work for any n ≥ 2. The line:
```
use [0, 1]
```
only provides a list of length 2, but I need to handle cases where n > 2 as well.

I need to generalize my construction. Perhaps I should create a list of length n instead of just assuming the base case will handle everything. The `simp` and `aesop` tactics might not be able to automatically construct this list for arbitrary n.

Let me reconsider this approach... What if I use a more general construction? I could use `List.replicate n 0` to create a list of n zeros. But then all elements would be the same, and the problem requires distinct integers.

Actually, I should think more carefully about what list would satisfy the divisibility condition. If I have integers a and b where a ≠ b, then I need (a + b) % (a - b) = 0.

For the general case, I might need to use a more sophisticated construction like arithmetic sequences or something based on factorials to ensure the divisibility property holds.

The issue with my current proof is that I'm not providing a general construction for n elements - I'm only handling n = 2. And even if `aesop` could somehow extend this, I haven't given it enough guidance on how to construct the larger list.

To fix this, I should:
1. Either provide an explicit construction for arbitrary n
2. Or use induction properly, showing how to extend from n to n+1 elements

I think I need to reconsider what list would work for this problem. Maybe something like powers of 2, or a list where consecutive elements have a specific relationship? The factorial approach mentioned in my comment might be promising - adding elements like a₁, a₂, ..., aₙ where each new element is constructed to maintain the divisibility relationship with all previous elements.

I should try constructing a list like [0, 1!, 2!, ..., (n-1)!] or something similar and verify that it satisfies the divisibility condition. This would give me distinct integers with potentially useful divisibility properties.
\end{lstlisting}

\section{RL Details} 
\label{sec:app_rl_details}

\begin{figure}[h]
     \centering
     \includegraphics[width=\linewidth]{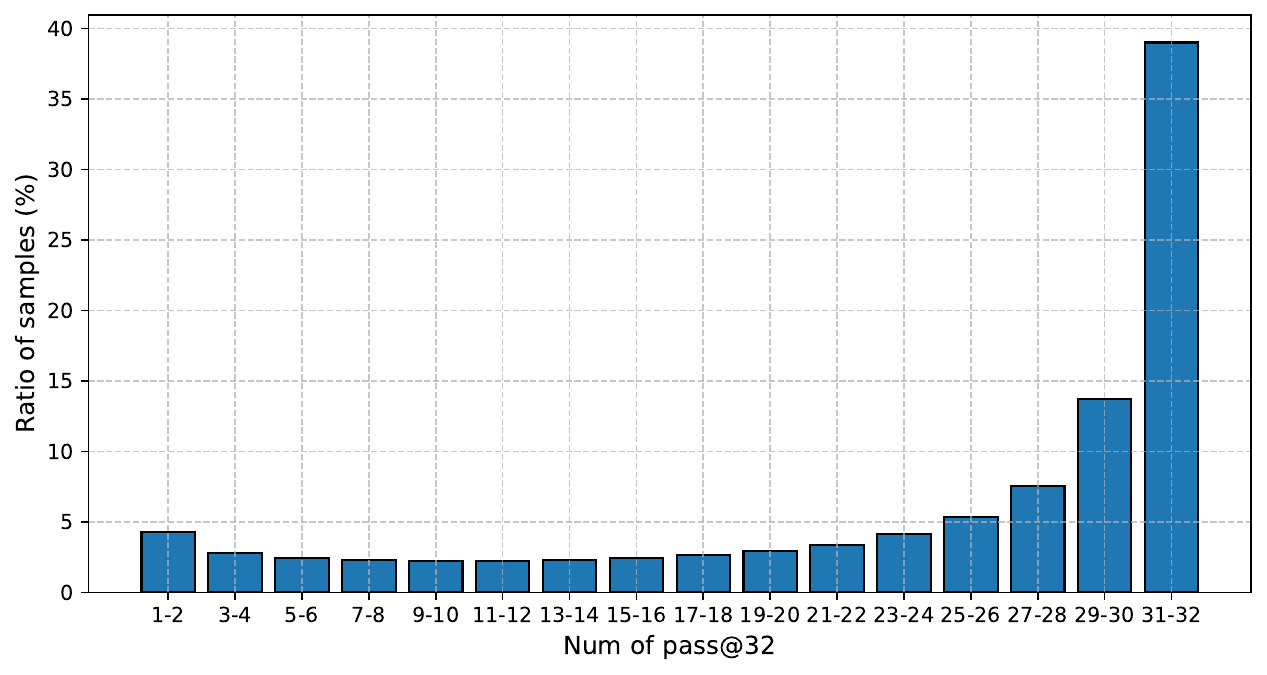}
     \caption{Distribution of valid proofs in pass@32. We calculate this distribution based on the \kylean-GD-SFT model. To build this set, we remain the formal statements who has at least sampled proofs are valid by Lean 4 compiler.}
     \label{fig:pass_dist}
\end{figure}

Figure~\ref{fig:pass_dist} shows the overall distribution of valid proofs in pass@32 testing. We collect the distribution base on our \kylean-GD-SFT model. 

\newpage

\end{CJK*}
\end{document}